\documentclass{article}

\usepackage{microtype}
\usepackage{graphicx}
\usepackage{subfigure}
\usepackage{amsmath,amssymb}
\usepackage{booktabs}
\usepackage[frozencache=true,cachedir=.]{minted}

\usepackage{listings}
\usepackage{color}
\definecolor{dkgreen}{rgb}{0,0.6,0}
\definecolor{gray}{rgb}{0.5,0.5,0.5}
\definecolor{mauve}{rgb}{0.58,0,0.82}
\DeclareMathOperator{\diag}{diag}

\usepackage{hyperref}

\usepackage[accepted]{icml2021}

\icmltitlerunning{Practical Laplace approximation using optimization second moments}

\begin{document}

\twocolumn[
\icmltitle{L2M: Practical posterior Laplace approximation \\ with optimization-driven second moment estimation}

\begin{icmlauthorlist}
	\icmlauthor{Christian S. Perone}{mlpoa}
	\icmlauthor{Roberto Pereira Silveira}{mlpoa}
	\icmlauthor{Thomas Paula}{mlpoa}
\end{icmlauthorlist}

\icmlaffiliation{mlpoa}{Machine Learning Porto Alegre. Porto Alegre/RS, Brazil}

\icmlcorrespondingauthor{Christian S. Perone}{christian.perone@gmail.com}

\icmlkeywords{Uncertainty estimation, Laplace, posterior approximation}

\vskip 0.3in
]



\printAffiliationsAndNotice{} %

\begin{abstract}
	Uncertainty quantification for deep neural networks has recently evolved through many techniques. In this work, we revisit Laplace approximation, a classical approach for posterior approximation that is computationally attractive. However, instead of computing the curvature matrix, we show that, under some regularity conditions, the Laplace approximation can be easily constructed using the gradient second moment. This quantity is already estimated by many exponential moving average variants of Adagrad such as Adam and RMSprop, but is traditionally discarded after training. We show that our method (L2M) does not require changes in models or optimization, can be implemented in a few lines of code to yield reasonable results, and it does not require any extra computational steps besides what is already being computed by optimizers, without introducing any new hyperparameter. We hope our method can open new research directions on using quantities already computed by optimizers for uncertainty estimation in deep neural networks.
\end{abstract}

\section{Introduction}
\label{introduction}
Uncertainty quantification is an important component of learning systems. It is often a required quantity for the deployment of deep learning models in real-world applications, especially when it involves decision making. Uncertainty is also widely used in many domains such as reinforcement learning~\cite{Deepmind}, active learning~\cite{Gal2017}, out-of-distribution detection~\cite{Lakshminarayanan2017}, and planning~\cite{Tigas2019}, to name a few.

Although many methods of uncertainty estimation for modern neural networks were recently developed, most of them rely on computationally expensive training and management procedures, sometimes of multiple models, with implementation complexity requiring specialized code for certain layer implementations or post-training tuning/estimation of hyperparameters in some cases.

Our contributions in this work are:
\begin{itemize}
\item We show that under some regularity conditions, a diagonal Laplace approximation can be constructed without computing anything besides what is already being computed by widely used optimizers;
\item We qualitatively compare this approximation with methods such as deep ensembles~\cite{Lakshminarayanan2017}, MC Dropout~\cite{GalDropout}, Hamiltonian Monte Carlo (HMC)~\cite{cobb2020scaling}, among others;
\item We also show that our approximation is orthogonal to methods such as ensembling~\cite{Lakshminarayanan2017} and does not require changing training procedures, estimating new quantities, or adding new hyperparameters.
\end{itemize}

To the best of our knowledge, it is the first time that the relationship between the raw second moment estimation in optimizers and uncertainty estimation is established. 

\subsection{Related Work}
The use of the Laplace method for posterior approximation in neural networks can be dated back to~\citet{WrayL.Buntine1991} and~\citet{MacKay1992}. Recently, it was also employed by~\citet{Ritter2018} in deep neural networks where a Kronecker factorization of the curvature matrix was employed. In~\citet{Kristiadi2020}, they also used Laplace approximation for the last layer of networks to overcome the confidence of ReLUs. In~\citet{Kirkpatrick2017}, a diagonal Laplace approximation was used to overcome catastrophic forgetting in neural networks.

Another line of work to estimate uncertainty in neural networks is through the use of ensembling methods as in~\citet{Lakshminarayanan2017} where multiple models are trained and then used to estimate the prediction variance. This method, together with random priors (randomly initialized and non-trainable networks) and bootstrapping was also employed in~\citet{Deepmind}.

Many variational approximations~\cite{Graves2011} were also employed for the posterior approximation, including an efficient and appealing method called MC Dropout by~\citet{GalDropout}, where dropout is left turned on during inference time.

Another method that was recently developed is called SWA-Gaussian (SWAG)~\cite{Maddox}, which relies on SWA~\cite{Izmailov2018} running averages for the first and second moment of parameters to build a Gaussian approximation of the posterior. The main difference of SWAG to our method is that it uses SGD and compute the moments of parameters, while our method re-uses optimizer second moment estimation of the gradients. Computationally, they have similar performance when using SWAG with SGD, but that is not the case if SWAG is used with other optimizers such as Adam~\cite{Kingma2015a}.

\section{Laplace approximation introduction}
The main idea of the Laplace approximation~\cite{Azevedo-Filho2013} can be stated as follows: locate the posterior mode and then construct a Gaussian distribution around the mode using as covariance matrix the inverse of the curvature around the mode.

The Laplace approximation can be derived from the second-order Taylor expansion around the mode (which can be found with optimization of a maximum a posterior $\theta_{\text{MAP}}$ estimator):
\begin{equation}
\log p(\theta \vert \mathcal{D}) \approx \log p(\theta_{\text{MAP}} \vert \mathcal{D}) - \frac{1}{2} (\theta - \theta_{\text{MAP}})^\intercal \tilde{H}(\theta - \theta_{\text{MAP}})
\end{equation}

Where $\tilde{H}$ is $\mathop{\mathbb{E}}[H]$, the expected Hessian of the negative log posterior. Given the property that we are expanding around $\theta_{\text{MAP}}$ where the gradient is zero, the first order term of the expansion becomes zero. By exponentiation we can then reach the functional form of the Gaussian posterior approximation:

\begin{equation}
\theta \sim \mathcal{N}(\theta_{\text{MAP}}, \tilde{H}^{-1})
\end{equation}

During inference, we can approximate the mean or any other statistic by taking $S$ Monte Carlo samples $\theta^{(s)}$ from this approximation:

\begin{equation}
p(\mathcal{D^*} \vert \mathcal{D}) = \int p(\mathcal{D}^* \vert \theta) p(\theta \vert \mathcal{D}) d\theta \approx \frac{1}{S} \sum^{S}_{S=1} p(\mathcal{D}^* \vert \theta^{(s)})
\end{equation}

Which gives us the posterior predictive distribution.

\section{Laplace approximation in neural networks}
The use of Laplace approximation of the posterior in neural networks can be dated back to~\citet{WrayL.Buntine1991} and~\citet{MacKay1992}. Only recently, Laplace approximation was used for larger modern deep neural networks~\cite{Ritter2018}, using a Kronecker factorization \cite{Martens2015a} of the curvature matrix.

Although the Kronecker factorization can yield a better approximation when compared to a diagonal approximation of the curvature matrix, the Kronecker factors still have to be computed on another step after training the network and can require specialized computation depending on the layers~\cite{Grosse2016} used in the network, making it difficult to use in practice. However, it is important to note that our method can be used together with KFAC~\cite{Martens2015a} instead of Adam, as the same approach described in this work can be used with the KFAC approximation instead of the raw diagonal second moment estimated by optimizers such as Adam~\cite{Kingma2015a}.

\section{Diagonal Laplace approximation using optimizer second moment}

\subsection{Diagonal Empirical Fisher estimation}
Many exponential moving average variants of Adagrad~\cite{Duchi2010} are popular in the deep learning community. RMSprop~\cite{Tieleman2012}, Adam~\cite{Kingma2015a}, and
Adadelta~\cite{Zeiler2012} are some examples. In this work, we will focus on Adam~\cite{Kingma2015a}, as it is one of the most used optimizers for deep neural networks.

The uncentered second moment of the gradients can be seen as a diagonal approximation of the Fisher Information Matrix (FIM) used in the Natural Gradient Descent~\cite{Amari1998}:

\begin{equation}
	\mathbf{F}_\theta = \mathop{\mathbb{E}}_{\substack{y \sim p_\theta (y \vert x) \\ x \sim p_{\text{data}}}} \left[ \nabla_\theta \log p_\theta(y \vert x) \, \nabla_\theta \log p_\theta(y \vert x)^\intercal \right] \,
\end{equation}

However, as noted by~\citet{Kunstner}, Adam optimizer does not use the expectation with $y \sim p_\theta (y \vert x)$ (with $y$ from the model distribution), but instead it uses the ground-truth labels. This matrix, which is often called ``empirical" Fisher (EF) \footnote{We use the same commonly nomenclature here as ``empirical" Fisher, however, as pointed by others~\cite{Thomas2019,Kunstner}, this term can be inaccurate.}, is often confused with the actual FIM. Also, in the preconditioning with the FIM, Adam employs a square root of the Fisher, another departure from the natural gradient.

\subsection{Construction of the Laplace approximation}
Under the condition where the model is realizable and there is enough data to recover the true parameters, then at the minimum, the EF converges to the FIM, as shown in~\cite{Kunstner}, given the consistency of the maximum likelihood estimator. Also, for optimizers using exponential moving averages of the gradient's second moment, the influence of gradients beyond a fixed window size becomes negligibly small~\cite{Reddi2018}.

Under the aforementioned regularity conditions regarding the convergence of the empirical Fisher to the $\mathbf{F}_\theta$ matrix, one can easily construct the posterior approximation using a diagonal Laplace approximation with the minima $\theta_\text{MAP}$ and the empirical Fisher diagonal $\diag(\mathbf{F}_\theta)$, quantities already estimated by many modern optimizers (and that are usually discarded after training):

\begin{equation}
    \theta \sim \mathcal{N}(\theta_\text{MAP}, \diag(\mathbf{F}_\theta)^{-1})
\end{equation}

The $\textbf{F}_\theta$ is connected to the Hessian through the Gauss-Newton matrix~\cite{Botev2017b}, which is an approximation of the Hessian where the second-order term is ignored, as it becomes small for models fitting data well. For exponential family distributions, the Gauss-Newton coincides with the Hessian, and as opposed to the Hessian -- that can be degenerate for neural networks -- the $\textbf{F}_\theta$ and the Gauss-Newton are guaranteed to be positive semi-definite (p.s.d.).

In practice, we also assume a Gaussian prior, that would be equivalent to the L$2$ term in regularized loss where $\theta \sim \mathcal{N}(0, \frac{1}{\lambda} \mathbb{I})$, $\lambda$ is the penalty term, and $\mathbb{I}$ denotes the identity matrix. This leads to the following approximation:

\begin{equation}
    \theta \sim \mathcal{N}(\theta_\text{MAP},\left[ \diag(\mathbf{F}_\theta) + \frac{1}{\lambda} + \epsilon \right]^{-1})
\end{equation}

Where $\epsilon$ is a small term to avoid division by zero. This approximation is very attractive in computational terms and simplicity, as it takes leverage of Adam's gradient second moment buffer (that are nowadays discarded) and uses the minima found by the optimizer as well. We call this approximation L2M (from \textbf{L}aplace \textbf{2}nd \textbf{M}oment).

It is also important to highlight that our method is orthogonal to methods such as deep ensembles~\cite{Lakshminarayanan2017}, since ensembles can be used to expand the mode coverage of the posterior while still using our method on each model of the ensemble, leveraging the local curvature information around individual modes of the posterior.

\subsection{Limitations}
Although our approximation is scalable and easy to implement, it also comes with limitations that are important to highlight. The Laplace approximation is a Taylor expansion around the MAP estimate and, in our case, its curvature is built with a diagonal approximation, which means that parameters are assumed to be independent. 

Also, the EF converges to the FIM only at the minimum~\cite{Kunstner}, and although it can have remarkable angle similarity~\cite{Thomas2019}, the ratio of traces from the EF and FIM can be quite different, which can lead to potential variance underestimation. However, up to a multiplicative constant, we can expect these two quantities to exhibit high similarity~\cite{Thomas2019}.

We note that priors other than Gaussian (which has L2 regularization equivalence) would require changes in our method. Nevertheless, for many applications, this approximation can be an excellent trade-off while providing reasonable uncertainty estimation with a simple, easy to implement, and scalable approach. 

\subsection{Note regarding the $\epsilon$ term}
The $\epsilon$ term, also introduced in Adam~\cite{Kingma2015a} to avoid division by zero, can also be seen as a damping term improving the conditioning of the Fisher or setting a trust region radius during optimization, as noted by~\citet{Choi2019}. In our case, $\epsilon$ can have an effect of reducing the posterior variance, as it is added to the Fisher's diagonal before the inversion.

\section{Evaluation}
Comparing different uncertainty estimation methods is a complex subject as there is no ground truth for uncertainty. Therefore, many works compare methods on the grounds of a qualitative evaluation on toy regression datasets or on out-of-distribution detection tasks.

In Figure~1, we show how our method compares to other methods on a toy regression dataset where $y \sim x^3 + \mathcal{N}(0, 3^2)$ as in~\citet{Ritter2018}. For this comparison, implemented in PyTorch~\cite{Paszke2019}, we employed a simple MLP with $2$ fully-connected layers, $40$ units and ReLU activation. We used an Adam's decoupled weight decay~\cite{Loshchilov2017} of $0.1$ with a learning rate of $0.1$ and trained it for $5,000$ epochs. For the predictions, we sampled $500$ posterior samples and then report the mean and $1$, $2$, and $3$ standard deviations above and below the mean.

\begin{figure*}[ht!]
	\vskip 0.1in
	\label{fig:qualitative-comp}
	\begin{center}  
		\includegraphics[width=\textwidth,trim=60 0 70 0, clip]{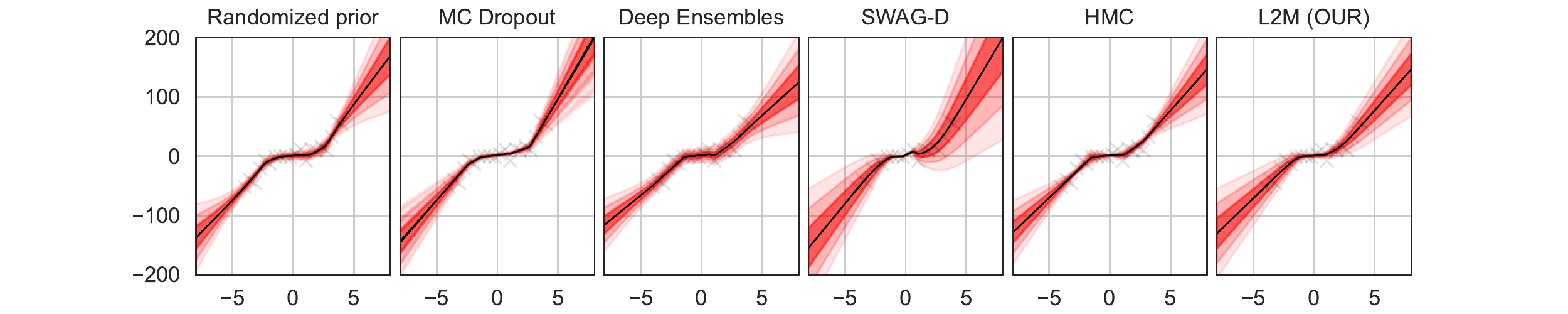}
		\caption{Qualitative comparison on a toy regression dataset. Red shades denotes 1, 2, and 3 $\sigma^2$. The black line shows the mean prediction and the crosses represents data points.
			\textbf{First panel} (\textit{from left to right}): Randomized prior functions~\cite{Deepmind}. \textbf{Second panel}: MC Dropout~\cite{GalDropout}. \textbf{Third panel}: Deep Ensembles~\cite{Lakshminarayanan2017}. \textbf{Fourth panel}: SWAG-Diagonal~\cite{Maddox}. \textbf{Fifth panel}: Hamiltonian Monte Carlo~\cite{cobb2020scaling}.  \textbf{Sixth panel}: L2M (our method).
		}
	\end{center}
	\vskip -0.2in
\end{figure*}

As we can clearly see in Figure~1, our method (L2M) performs on par with other methods, i.e., uncertainty is high in regions with no data points and low in regions with several data points. The method here presented does not need lots of hyperparameters to tune and does not need post-processing steps in order to get to these results, while still being fast to get the posterior predictive distribution.

We leave the evaluation of the integration of our method with deep ensembles~\cite{Lakshminarayanan2017} for future work, as our method can be used on each model of the ensemble to approximate the local curvature around the mode, potentially leading to better uncertainty estimates. The application of our method to classification problems is also left for future work.

\section{Conclusion}
In this paper, we presented L2M, a method for uncertainty quantification for deep neural networks. We revisited Laplace approximation, a classical approach for posterior approximation that is computationally attractive, and showed that we can construct a posterior approximation with the gradient raw second moment which is already computed by many exponential moving average variants of Adagrad~\cite{Duchi2010}. To the best of our knowledge, it is the first time that this relationship between the raw second moment estimation in optimizers and uncertainty estimation was established.

We qualitatively compared the approximations produced by our method with other commonly used methods such as MC Dropout~\cite{GalDropout}, Deep Ensembles~\cite{Lakshminarayanan2017}, among others. Our results are encouraging and open new research directions towards scalable approaches leveraging raw second moment estimation in optimizers for uncertainty estimation.

L2M has some limitations though. First, the curvature is built with a diagonal approximation, which means that parameters are assumed to be independent. Second, we rely on some regularity conditions such as the convergence of the EF (``Empirical" Fisher) to the FIM (Fisher Information Matrix). Third, priors other than Gaussian would require changes in our method, as the prior equivalence with the Gaussian prior wouldn't hold anymore. And last, we tested our approach only in a toy regression dataset, which means it would need to be further expanded to other larger datasets, network architectures and tasks.

For future work, we plan to conduct experiments on larger datasets. We also plan to expand this method to handle general classification and language modelling tasks with larger networks. Finally, we aim to explore the use of this method for out-of-distribution problems and incorporate it into other techniques such as deep ensembles~\cite{Lakshminarayanan2017}.

\section*{Software and Data}
A PyTorch~\cite{Paszke2019} implementation of our method can be found at the following repository below:

\textbf{\url{https://github.com/anonymous/L2M}}

\bibliography{example_paper}
\bibliographystyle{icml2021}

\newpage
\appendix
\section*{Appendix}
\subsection*{Appendix A - L2M approximation pseudo-code}
This appendix shows a pseudo-code implementation of our method:
\begin{minted}[
frame=lines,
framesep=3.0mm,
baselinestretch=1.2,
fontsize=\footnotesize,
]{python}
# Trained model and optimizer
model = (...)
optimizer = (...)

# Build L2M approximation
grad_second_moment = optimizer.second_moment()
prior_variance = 1.0/weight_decay
fisher_diag = grad_second_moment + \
prior_variance + eps
inv_fisher_diag = 1.0/fisher_diag
l2m_approximation = Normal(model.weights,
sqrt(inv_fisher_diag))

# Multiple posterior samples for inference
outputs = [] # predictive distribution
for s in S:
model.weights = l2m_approximation.sample()
output = model(input)
outputs.append(output)
\end{minted}

\end{document}